\documentclass{article}

\usepackage{times}

\usepackage{amssymb,amsmath}
\usepackage{epsfig,graphicx}
\usepackage{verbatim}
\usepackage{amsfonts}
\usepackage{times}
\usepackage{helvet}
\usepackage{courier}
\usepackage{graphicx}
\usepackage{graphicx}
\usepackage{wrapfig}

\usepackage{amsmath}
\usepackage{url}
\usepackage{verbatim}

\usepackage{multirow}
\usepackage{amssymb}

\long\def\comment#1{}

\newcommand{\D}{\:\: |\:\:}

\newtheorem{examp}{Example}
\newtheorem{srss}{Structured Rule System}

\title{From First-Order Logic to Assertional Logic}

\author{Yi Zhou}

\date{}

\begin{document}

\maketitle

\begin{abstract}
First-Order Logic (FOL) is widely regarded as one of the most
important foundations for knowledge representation. Nevertheless, in
this paper, we argue that FOL has several critical issues for this
purpose. Instead, we propose an alternative called assertional
logic, in which all syntactic objects are categorized as set
theoretic constructs including individuals, concepts and operators,
and all kinds of knowledge are formalized by equality assertions. We
first present a primitive form of assertional logic that uses
minimal assumed knowledge and constructs. Then, we show how to
extend it by definitions, which are special kinds of knowledge,
i.e., assertions. We argue that assertional logic, although simpler,
is more expressive and extensible than FOL. As a case study, we show
how assertional logic can be used to unify logic and probability,
and more building blocks in AI.
\end{abstract}

\section{Introduction}\label{sec-intro}

Classical First-Order Logic (FOL) is widely regarded as one of the
most important foundations of symbolic AI. FOL plays a central role
in the field of Knowledge Representation and Reasoning (KR). Many of
its fragments (such as propositional logic, modal and epistemic
logic, description logics), extensions (such as second-order logic,
situation calculus and first-order probabilistic logic) and
variants (such as Datalog and first-order answer set programming)
have been extensively studied in the literature
\cite{Brachman04,Harmelen08}.

Nevertheless, AI researchers have pointed out several issues
regarding using FOL for knowledge representation and reasoning,
mostly from the reasoning point of view. First, FOL is
computationally very difficult. Reasoning about FOL is a well-known
undecidable problem. Also, FOL is monotonic in the sense that adding
new knowledge into a first-order knowledge base always results in
more consequences. However, human reasoning is sometimes
nonmonotonic.

In this paper, we argue that FOL also has some critical issues from
the knowledge representation point of view. First of all, although
FOL is considered natural for well-trained logicians, it is not
simple and flexible enough for knowledge engineers with less
training. One reason is the distinction and hierarchy between term
level, predicate level and formula level. From my own experience as
a teacher in this subject, although strongly emphasized in the
classes, many students failed to understand why a predicate or a
formula cannot be in the scope of a function. Another reason is the
notion of free occurrences of variables. For instance, it is not
easily understandable for many students why the GEN inference rule
has to enforce the variable occurrence restrictions. Last but not
least, arbitrary nesting also raises issues. Although natural from a
mathematical point of view, a nested formula, e.g., $(x \lor \neg (y
\land z)) \land (\neg y \lor \lnot x)$ is hard to be understood and
used.

Secondly, FOL has limitations in terms of expressive power. FOL
cannot quantify over predicates/functions. This can be addressed by
extending FOL into high-order logic. Nevertheless, high-order logic
still cannot quantify over formulas. As a consequence, FOL and
high-order logic are not able to represent an axiom or an inference
rule in logic, such as {\it Modus Ponens}. Flexible quantification
beyond the term level is needed in applications. As an example, in
automated solving mathematical problems, we often use proof by
induction. To represent this, we need to state that for some
statement $P$ with a number parameter, if that $P$ holds for all
numbers less than $k$ implies that $P$ holds for the number $k$ as
well, then $P$ holds for all natural numbers. Here, $P$ is a
statement at a formula level, possibly with complex sub-statements
within itself. Hence, in order to represent and use proof by
induction, we need to quantify over $P$ that is at a formula level.

Thirdly, FOL is hard to be extended with new building blocks. FOL
itself cannot formalize some important AI notions including
probability, actions, time etc., which are needed in a wide range of
applications. For this purpose, AI researchers have made significant
progresses on extending FOL with these notions separately, such as
first-order probabilistic logic \cite{Bacchus1991,zbMATH04191596},
situation calculus \cite{levesque1991,Lin08}, CTL \cite{Clarke1981}
etc. Each is a challenging task in the sense that it has to
completely re-define the syntax as well as the semantics. However,
combing these notions together, even several of them, seems an
extremely difficult task. Moreover, there are many more building
blocks to be incorporated in applications. For instance, consider
task planning for home service robots \cite{Keller10}. It is
necessary to represent actions, probability, time and more building
blocks such as preferences altogether at the same time.

To address these issues, we propose assertional logic, in which all
syntactic objects are categorized as set theoretic constructs
including individuals, concepts and operators, and all kinds of
knowledge are uniformly formalized by equality assertions of the
form $a=b$, where $a$ and $b$ are either atomic individuals or
compound individuals. Semantically, individuals, concepts and
operators are interpreted as elements, sets and functions
respectively in set theory, and knowledge of the form $a=b$ means
that the two individuals $a$ and $b$ are referring to the same
element.

We first present the primitive form of assertional logic that uses
minimal assumed knowledge and primitive constructs. Then, we show
how to extend it with more building blocks by definitions, which are
special kinds of knowledge, i.e., assertions used to define new
individuals, concepts and operators. Once these new syntactic
objects are defined, they can be used as a basis to define more. As
an example, we show how to define multi-assertions by using
Cartesian product, and nested assertions by using multi-assertions.

We show that assertional logic, although simpler, is more expressive
and extensible than FOL. As a case study, we show how to extend
assertional logic for unifying logic and probability, and more
important AI building blocks including time. Note that our intention
is not to reinvent the wheel of these building blocks but to borrow
existing excellent work on formalizing these building blocks
separately and to assemble them within one framework (i.e., assertional
logic) so that they can live happily ever after.

\section{Meta Language and Prior Knowledge}

One cannot build something from nothing. Hence, in order to
establish assertional logic, we need some basic knowledge. Of
course, for the purpose of explanation, we need an informal meta
language whose syntax and semantics are pre-assumed. As usual, we
use a natural language such as English. Nevertheless, this meta
language is used merely for explanation and it should not affect the
syntax as well as the semantics of anything defined formally.

Only a meta level explanation language is not enough. Other than
this, we also need some core objects and knowledge, whose syntax and
semantics are pre-assumed as well. These are called {\em prior
objects} and {\em prior knowledge}. For instance, when defining real
numbers, we need some prior knowledge about natural numbers; when
defining probability, we need some prior knowledge about real
numbers.

In assertional logic, we always treat the equality symbol ``$=$" as
a prior object. There are some prior knowledge associated with the
equality symbol. For instance, ``$=$" is an equivalence relation
satisfying reflexivity, symmetricity, and transitivity. Also, ``$=$"
satisfies the general substitution property, that is, if $a=b$, then
$a$ can be used to replace $b$ anywhere. Other than the equality
symbol, we also assume some prior objects and their associated prior
knowledge in set theory \cite{Halmos60}, including set operators
such as set union and Cartesian product, Boolean values, set builder
notations and natural numbers.

\section{Assertional Logic: the Primitive Form}\label{sec-pf}

In this section, we present the primitive form of assertional logic.
As the goal of assertional logic is to syntactically represent
knowledge in application domains, there are two essential tasks,
i.e., how to capture the syntax of the domain and how to represent
knowledge in it.

\subsection{Capturing the syntax}

Given an application domain, a {\em syntactic structure} ({\em
structure} for short if clear from the context) of the domain is a
triple $\langle \mathcal{I}, \mathcal{C}, \mathcal{O} \rangle$,
where $\mathcal{I}$ is a collection of {\em individuals},
representing objects in the domain, $\mathcal{C}$ a collection of
{\em concepts}, representing groups of objects sharing something in
common and $\mathcal{O}$ a collection of {\em operators},
representing relationships and connections among individuals and
concepts. Concepts and operators can be nested and considered as
individuals as well. If needed, we can have concepts of concepts,
concepts of operators, concepts of concepts of operators and so on.

An operator could be multi-ary, that is, it maps a tuple of
individuals into a single one. Each multi-ary operator $O$ is
associated with a {\em domain} of the form $(C_1, \dots, C_n)$,
representing all possible values that the operator $O$ can operate
on, where $C_i, 1 \le i \le n$, is a concept. We call $n$ the {\em
arity} of $O$. For a tuple $(a_1,\dots,a_n)$ matching the domain of
an operator $O$, i.e., $a_i \in C_i, 1 \le i \le n$, $O$ maps
$(a_1,\dots,a_n)$ into an individual, denoted by $O(a_1,\dots,a_n)$.

\comment{ We also use $O(C_1,\dots,C_n)$ to denote the set
$\{O(a_1,\dots,a_n) \D a_i \in C_i\}$, called the {\em range} of the
operator $O$.
 \footnote{Note that in set theory, a tuple of sets is a
Cartesian product of some sets, which itself is a set as well.
Therefore, multi-ary operators can essentially be viewed as unary.}
}

Operators are similar to functions in FOL but differs in two
essential ways. First, operators are many-sorted as $C_1, \dots,
C_n$ could be different concepts. More importantly, $C_1, \dots,
C_n$ could be high-order constructs, e.g., concepts of concepts,
concepts of operators.

\comment{For instance, consider a family relationship domain, in
which $Alice$ and $Bob$ are individuals, $Human$, $Woman$ and
$Female$ are concepts and $Father$, $Mother$ and $Aunt$ are
operators etc.

 For instance, consider the arithmetic domain, in which
$0$, $1$, $2$, etc., are individuals; the set $\mathbb{N}$ of
natural numbers is a concept; the successor operator $Succ$ and the
add operators $Add$ are operators.}

\comment{ For convenience, if $O$ is unary, we sometimes use $a.O$
($C.O$) to denote $O(a)$ ($O(C)$), where $a \in \mathcal{I}$ and $C
\in \mathcal{C}$. If $O$ is binary, we sometimes use $a\:O\:b$ ($A\:
O \: B$) to denote $O(a,b)$ ($O(A,B)$), where $a, b \in \mathcal{I}$
and $A,B \in \mathcal{C}$. If the range of an operator $O$ is
Boolean, we sometimes use $O(a_1,\dots,a_n)$ to denote
$O(a_1,\dots,a_n)=\top$.}

\comment{ We use pre-given, established syntactic objects
(individuals, concepts and operators) to define new ones. The former
is said to be {\em prior} while the latter is said to be {\em
posterior}. Prior objects are assumed knowledge while posterior
objects are knowledge to be studied. Prior and posterior objects are
relative with respect to a particular application domain. For
instance, when we use the successor operator to define the add
operator, the former is prior whilst the latter is posterior.
Nevertheless, when we use the add operator to define something else,
it becomes prior. }

\subsection{Representing knowledge}

Let $\langle \mathcal{I}, \mathcal{C}, \mathcal{O} \rangle$ be a
syntactic structure. A {\em term} is an individual, either an atomic
individual $a\in \mathcal{I}$ or the result $O(a_1,\dots,a_n)$ of an
operator $O$ operating on some individuals $a_1,\dots,a_n$. We also
call the latter {\em compound individuals}.

\comment{ A term is said to be {\em prior} if all individuals and
operators in it are prior. Otherwise, it is said to be {\em
posterior}.}

An {\em assertion} is of the form
\begin{equation}\label{form-assertion}
a=b,
\end{equation}
where $a$ and $b$ are two terms. Intuitively, an assertion of the
form (\ref{form-assertion}) is a piece of knowledge in the
application domain, claiming that the left and the right side refer
to the same object. A {\em knowledge base} is a set of assertions.
Terms and assertions can be considered as individuals as well.

\comment{Here, $=$ is the equality binary operator among
individuals, and it is always considered as a prior knowledge. In
this sense, $a=b$ can be understood in alternative way that $=(a,b)$
is true.

 For instance, in the
family relationship domain, $Father(Alice)=Bob$,
$Father(Alice)=Uncle(Bob)$ are assertions.}

\comment{Assertion consistent, deterministic}

Similar to concepts that group individuals, we use schemas to group
terms and assertions. A {\em schema term} is either an atomic
concept $C\in \mathcal{C}$ or of the form $O(C_1,\dots, C_n)$, where
$C_i, 1 \le i \le n$ are concepts. Essentially, a schema term
represents a set of terms, in which every concept is grounded by a
corresponding individual. That is, $O(C_1,\dots, C_n)$ is the
collection $\{O(a_1,\dots, a_n)\}$, where $a_i \in C_i, 1 \le i \le
n$ are individuals. Then, a {\em schema assertion} is of the same
form as form (\ref{form-assertion}) except that terms can be
replaced by schema terms. Similarly, a schema assertion represents a
set of assertions.

We say that a schema term/assertion {\em mentions} a set
$\{C_1,\dots,C_n\}$ of concepts if $C_1,\dots,C_n$ occur in it, and
{\em only mentions} if $\{C_1,\dots,C_n\}$ contains all concepts
mentioned in it. Note that it could be the case that two or more
different individuals are referring to the same concept $C$ in
schema terms and assertions. In this case, we need to use different
{\em copies} of $C$, denoted by $C^1,C^2,\dots$, to distinguish
them. For instance, all assertions $x=y$, where $x$ and $y$ are
human, are captured by the schema assertion $Human^1=Human^2$. On
the other side, in a schema, the same copy of a concept $C$ can only
refer to the same individual. For instance, $Human=Human$ is the set
of all assertions of the form $x=x$, where $x \in Human$.

\subsection{The semantics}

We propose a set theoretic semantics for assertional logic. Since we
assume set theory as the prior knowledge, in the semantics, we
freely use those individuals (e.g., the empty set), concepts (e.g.,
the set of all natural numbers) and operators (e.g., the set union
operator) without explanation.

An {\em interpretation} (also called a {\em possible world}) is a
pair $\langle \Delta, .^{I} \rangle$, where $\Delta$ is a domain of
elements, and $.^{I}$ is a mapping function that admits all prior
knowledge, and maps each individual into a domain element in
$\Delta$, each concept into a set in $\Delta$ and each $n$-ary
operator into an $n$-ary function in $\Delta$. The mapping function
$.^{I}$ is generalized for terms by mapping $O(a_1,\dots,a_n)$ to
$O^I(a_1^I,\dots,a_n^I)$. Similar to terms and assertions,
interpretations can also be considered as individuals to be studied.

It is important to emphasize that an interpretation has to admit all
prior knowledge. For instance, since we assume set theory, suppose
that an interpretation maps two individuals $x$ and $y$ as the same
element $a$ in the domain, then the concepts $\{x\}$ and $\{y\}$
must be interpreted as $\{a\}$, and $x=y$ must be interpreted as
$a=a$.

Let $I$ be an interpretation and $a=b$ an assertion. We say that $I$
is a {\em model} of $a=b$, denoted by $I \models a=b$ iff
$.^{I}(a)=.^{I}(b)$, also written $a^I=b^I$. Let $KB$ be a knowledge
base. We say that $I$ is a model of $KB$, denoted by $I \models KB$,
iff $I$ is a model of all assertions in $KB$. We say that an
assertion $A$ is a {\em property} of $KB$, denoted by $KB \models
A$, iff all models of $KB$ are also models of $A$. In particular, we
say that an assertion $A$ is a {\em tautology} iff it is modeled by
all interpretations.

Since we assume set theory as our prior knowledge, we directly
borrow some set theoretic constructs. For instance, we can use $\cup
(C_1, C_2)$ (also written as $C_1 \cup C_2$) to denote a new concept
that unions two concepts $C_1$ and $C_2$. Applying this to
assertions, we can see that assertions of the primitive form
(\ref{form-assertion}) can indeed represent many important features
in knowledge representation. For instance, the {\em membership
assertion}, stating that an individual $a$ is an instance of a
concept $C$, is the following assertion $\in(a,C)=\top$ (also
written as $a \in C$). The {\em containment assertion}, stating that
a concept $C_1$ is contained by another concept $C_2$, is the
following assertion $\subseteq(C_1,C_2)=\top$ (also written as $C_1
\subseteq C_2$). The {\em range declaration}, stating that the range
of an operator $O$ operating on some concepts $C_1,\dots,C_n$ equals
to another concept $C$, is the following assertion
$O(C_1,\dots,C_n)=C$.

\section{Extensibility via Definitions}

As argued in the introduction section, extensibility is a critical
issue for knowledge representation. In assertional logic, we use
{\em definitions} for this purpose. Definitions are (schema)
assertions used to define new syntactic objects (including
individuals, concepts and operators) based on existing ones. Once
these new syntactic objects are defined, they can be used to define
more. Note that definitions are nothing extra but special kinds of
knowledge (i.e. assertions).

We start with defining new individuals. An individual definition is
an assertion of the form
\begin{equation}\label{form-individual-definition}
a=t,
\end{equation}
where $a$ is an atomic individual and $t$ is a term. Here, $a$ is the
individual to be defined. This assertion claims that the left side
$a$ is defined as the right side $t$. For instance, $0=\emptyset$
means that the individual $0$ is defined as the empty set.

Defining new operators is similar to defining new individuals except
that we use schema assertions instead. Let $O$ be an operator to be
defined and $(C_1,\dots,C_n)$ its domain. An operator definition is
a schema assertion of the form
\begin{equation}\label{form-operator-definition}
O(C_1,\dots,C_n)=T,
\end{equation}
where $T$ is a schema term that mentions concepts only from
$C_1,\dots,C_n$.

\comment{It could be the case that $T$ only mentions some of $C_1,
\dots, C_n$. Note that if $C_1,\dots,C_n$ refer to the same concept,
we need to use different copies. }

Since a schema assertion represents a set of assertions,
essentially, an operator definition of the form
(\ref{form-operator-definition}) defines the operator $O$ by
defining the value of $O(a_1,\dots,a_n)$ one-by-one, where $a_i \in
C_i, 1 \le i \le n$. For instance, for defining the successor
operator $Succ$, we can use the schema assertion
$Succ(\mathbb{N})=\{\mathbb{N}, \{\mathbb{N}\}\}$, meaning that, for
every natural number $n$, the successor of $n$, is defined as
$\{n,\{n\}\}$.

Defining new concepts is somewhat different. As concepts are
essentially sets, we directly borrow set theory notations for this
purpose. There are four ways to define a new concept.

\noindent {\bf Enumeration} Let $a_1,\dots,a_n$ be $n$ individuals.
Then, the collection $\{a_1,\dots, a_n\}$ is a concept, written as
\begin{equation}
C=\{a_1,\dots, a_n\}.
\end{equation}
For instance, we can define the concept $Digits$ by
$Digits=\{0,1,2,3,4,5,6,7,8,9\}$.

\noindent {\bf Operation} Let $C_1$ and $C_2$ be two concepts. Then,
$C_1 \cup C_2$ (the union of $C_1$ and $C_2$), $C_1 \cap C_2$ (the
intersection of $C_1$ and $C_2$), $C_1 \setminus C_2$ (the
difference of $C_1$ and $C_2$), $C_1 \times C_2$ (the Cartesian
product of $C_1$ and $C_2$), $2^{C_1} $ (the power set of $C_1$) are
concepts. Operation can be written by assertions as well. For
instance, the following assertion
\begin{equation}
C=C_1 \cup C_2
\end{equation}
states that the concept $C$ is defined as the union of $C_1$ and
$C_2$. As an example, one can define the concept $Man$ by
$Man=Human\cap Male$.

\noindent {\bf Restricted Comprehension} Let $C$ be a concept and
$A(C)$ a schema assertion that only mentions concept $C$. Then,
individuals in $C$ satisfying $A$, denoted by $\{x\in C | A(x)\}$
(or simply $C|A(C)$), form a concept, written as
\begin{equation}
C'=C|A(C).
\end{equation}
For instance, we can define the concept $Male$ by $Male =\{Animal \D
Sex(Animal)=male\}$, meaning that $Male$ consists of all animals
whose sexes are male.

\noindent {\bf Replacement} Let $O$ be an operator and $C$ a concept
on which $O$ is well defined. Then, the individuals mapped from $C$
by $O$, denoted by $\{O(x) \D  x \in C\}$ (or simply $O(C)$), form a
concept, written as
\begin{equation}
C'=O(C).
\end{equation}
For instance, we can define the concept $Parents$ by
$Parents=ParentOf(Human)$, meaning that it consists of all
individuals who is a $ParentOf$ some human.

Definitions can be incremental. We may define some syntactic objects
first. Once defined, they can be used to define more. One can always
continue with this incremental process. For instance, in arithmetic,
we define the successor operator first. Once defined, it can be used
to define the add operator, which is further served as a basis to
define more.

\comment{ For clarity, we use the symbol ``$::=$" to replace ``$=$"
for definitions. }

Since terms and assertions can be considered as individuals, we can
define new type of terms and assertions by definitions. As an
example, we extend assertions of the form (\ref{form-assertion})
into multi-assertions by using Cartesian product. We first define
multi-assertions of a fixed number of assertions. Given a number
$n$, we define a new operator $M_n$ with arity $n$ by the following
schema assertion: \small
\begin{equation*}
M_n(C_1=D_1,\dots,C_n=D_n)=(C_1,\dots,C_n)=(D_1,\dots,D_n),
\end{equation*}
\normalsize where $C_i, D_i, 1 \le i \le n$, are concepts of terms.
Notice that, $(C_1,\dots,C_n)=(D_1,\dots,D_n)$ is a single assertion
of the form (\ref{form-assertion}). In this sense, an $n$-ary
multi-assertion is just a syntax sugar. Then, we define the concept
of multi-assertions:
\[
Multi-Assertion = \bigcup_{1 \le i \le \infty}
M_i(\mathcal{A}^1,\dots,\mathcal{A}^i),
\]
where $\mathcal{A}^1,\dots,\mathcal{A}^i$ are $i$ copies of standard
assertions. For convenience, we use $ Assertion_1, \dots,
Assertion_n$ to denote an $n$-ary multi-assertion. Once
multi-assertion is defined, it can be used to define more syntactic
objects.

As an example, we use multi-assertion to define nested assertions.
We first define nested terms as follows:
\begin{eqnarray*}
Nested-Term & = & Term \cup N-Term \\
N-Term & = & Op(Nested-Term).
\end{eqnarray*}
Then, nested assertions can be defined as
\[
Nested-Assertion=Nested-Term = Nested-Term.
\]

\comment{ Nested assertions can be represented by non-nested
multi-assertions by introducing new individuals. Whenever a result
of nested term is used, we introduce a new individual to replace it
and claim that this new individual is defined as the nested term.
That is, for every nested term
$Op(a_1,\dots,Op'(b_1,\dots,b_m),\dots,a_m)$ occurred in a nested
assertion, we introduce a new atomic individual $a'$; replace the
above term with $Op(a_1,\dots,a',\dots,a_m)$ and add a new assertion
$a'=Op'(b_1,\dots,b_m)$. For instance, the nested assertion
$Op(a,Op(b,Op'(c)))=Op'(d)$ is defined as $Op(a,x)=Op'(d),
x=Op(b,y), y=Op'(c)$, where $x$ and $y$ are new individuals. In this
sense, nested assertion is essentially a multi-assertion, which can
be represented as a single assertion. Therefore, nested assertion is
a syntactic sugar of the primitive form as well.}

Again, once nested assertion is defined, it can be used as basis to
define more, so on and so forth. Using nested assertions can
simplify the representation task. However, one cannot overuse it
since, essentially, every use of a nested term introduces a new
individual.

\section{Embedding FOL into Assertional
Logic}\label{sec-logic}

In the previous section, we show how to extend assertions of the
primitive form (\ref{form-assertion}) into multi-assertions and
nested assertions. In this section, we continue with this task to
show how to define more complex forms of assertions with logic
connectives, including propositional connectives and quantifiers.

We start with the propositional case. Let $\mathcal{A}$ be the
concept of nested assertions. We introduce a number of operators
over $\mathcal{A}$ in assertional logic, including
$\neg(\mathcal{A})$ (for {\em negation}),
$\land(\mathcal{A}^1,\mathcal{A}^2)$ (for {\em conjunction}),
$\lor(\mathcal{A}^1,\mathcal{A}^2)$ (for {\em disjunction}) and $\to
(\mathcal{A}^1,\mathcal{A}^2)$ (for {\em implication}).

There could be different ways to define these operators in
assertional logic. Let $a=a'$ and $b=b'$ be two (nested) assertions.
The propositional connectives are defined as follows: \small
\begin{align*}
\neg(a=a')& = && \hspace{-.1in} \{a\}\cap\{a'\}=\emptyset \\
\land(a=a',b=b') & = && \hspace{-.1in} (\{a\}\cap\{a'\}) \cup
(\{b\}\cap\{b'\})=\{a,a',b,b'\} \\
\lor(a=a',b=b') & = && \hspace{-.1in} (\{a\}\cap\{a'\}) \cup
(\{b\}\cap\{b'\}) \ne \emptyset  \\
\to (a=a',b=b') & = && \hspace{-.1in} (\{a,a'\} \setminus
\{a\}\cap\{a'\})\cup (\{b\}\cap\{b'\}) \ne \emptyset,
\end{align*}
\normalsize where $a\ne a'$ is used to also denote $\neg(a=a')$. One
can observe that the ranges of all logic operators are nested
assertions. Hence, similar to multi-assertion and nested assertion,
propositional logic operators are syntactic sugar as well.

Now we consider to define operators for quantifiers, including
$\forall$ (for the {\em universal} quantifier) and $\exists$ (for
the {\em existential} quantifier). The domain of quantifiers is a
pair $(C,A(C))$, where $C$ is a concept and $A(C)$ is a schema
assertion that only mentions $C$.

The quantifiers are defines as follows:
\begin{eqnarray}
\forall(C,A(C)) & = & C|A(C)=C \label{universal-quantifier}\\
\exists(C,A(C)) & = & C|A(C) \ne \emptyset \label{exist-quantifier}
\end{eqnarray}
Intuitively, $\forall(C,A(C))$ is true iff those individuals $x$ in
$C$ such that $A(x)$ holds equals to the concept $C$ itself, that
is, for all individuals $x$ in $C$, $A(x)$ holds; $\exists(C,A(C))$
is true iff those individuals $x$ in $C$ such that $A(x)$ holds does
not equal to the empty set, that is, there exists at least one
individual $x$ in $C$ such that $A(x)$ holds. We can see that the
ranges of quantifiers are nested assertions as well. In this sense,
quantifiers are also syntactic sugar of the primitive form.

Note that quantifiers defined here are ranging from an arbitrary
concept $C$. If $C$ is a concept of all atomic individuals and all
quantifiers range from the same concept $C$, then these quantifiers
are first-order. Nevertheless, the concepts could be different. In
this case, we have many-sorted FOL. Moreover, $C$ could be complex
concepts, e.g., a concept of all possible concepts. In this case, we
have monadic second-order logic. Yet $C$ could be many more, e.g., a
concept of assertions, a concept of concepts of terms etc. In this
sense, the quantifiers become high-order. Finally, the biggest
difference is that $C$ can even be a concept of assertions so that
quantifiers in assertional logic can quantify over assertions
(corresponding to formulas in classical logics), while this cannot
be done in classical logics including high-order logic.

It can be verified that all tautologies in propositional logic and
FOL (e.g., De-Morgan's laws) are also tautologies in assertional
logic. For space reasons, we leave the theorems and their proofs to
a full version of this paper.

\comment{ A problem arises whether there is cyclic definition as we
assume first-order logic as our prior knowledge. Nevertheless,
although playing similar roles, operators (over assertions) defined
in assertional logic are considered to be different from logic
connectives (over propositions/formulas) since they are on a
different layer of definition. The main motivation is for the
purpose of extensibility, i.e., by embedding classical logic
connectives into operators in assertional logic, we can easily
extend it with more components and building blocks including
probability.}

\section{Incorporating Probability and More}\label{sec-probability}

Probability is another important building block for knowledge
representation. In the last several decades, with the development of
uncertainty in artificial intelligence, a number of influential
approaches
\cite{Bacchus1991,Gaifman64,zbMATH03871328,Milch:2006:PMU:1292992,Pearl1988,Richardson06}
have been developed, and important applications have been found in
machine learning, natural language processing etc.

Normally, to incorporate probability in logic, one has to complete
redefine the whole semantics since the integrations between
probability and logic connectives and quantifiers are complicated.
In this section, we show how this can be easily done in assertional
logic. The key point is, although the interactions between logic and
probability are complicated, their interactions with assertions of
the basic form (\ref{form-assertion}) is relatively simple. As shown
in the previous section, the interactions between logic and
assertions can be defined by a few lines. In this section, following
Gaifman's idea \cite{Gaifman64}, we show that this is indeed the
case for integrating assertions with probability as well. Then, the
interactions between logic and probability will be automatically
established via assertions.

\subsection{Integrating assertions with
probability}\label{sec-define-probability}

Since operations over real numbers are involved in defining
probability, we need to assume a theory of real number as our prior
knowledge.

Gaifman \cite{Gaifman64} proposed to define the probability of a
logic sentence by the sum of the probabilities of the possible
worlds satisfying it. Following this idea, in assertional logic, we
introduce an operator $Pr$ (for probability) over the concept
$\mathcal{A}$ of assertions. The range of $Pr$ is the concept of
real numbers. For each possible world $w$, we assign an associated
weight $W_w$, which is a positive real number. Then, for an
assertion $A$, the probability of $A$, denoted by $Pr(A)$, is define
by the following schema assertion:
\begin{equation}\label{def-probability}
Pr(A)= \frac{\Sigma_{w, w \models A} \:\: W_w}{\Sigma_{w} \: \:W_w}.
\end{equation}
This definition defines the interactions between probability and
assertions. In case that there are a number of infinite worlds, we
need to use measure theory. Nevertheless, this is beyond the scope
of our paper.

\comment{Nevertheless, this is beyond the scope of our paper, which
focuses on how to use assertional logic for extensible knowledge
modeling.}

Once we have defined the probability $Pr(A)$ of an assertion $A$ as
a real number, we can directly use it inside other assertions. In
this sense, $Pr(A)=0.5$, $Pr(A)\ge 0.3$, $Pr(A) \ge Pr(\forall
(C,B(C)))-0.3$, $Pr(A) \times 0.6 \ge 0.4$ and $Pr(Pr(A) \ge 0.3)
\ge 0.3$ are all valid assertions. We are able to vefiry some
properties about probability, for instance, Kolmogorov's first and
second probability axioms.

\comment{that is, 1 for all assertions, $ Pr(A) \ge 0$, and 2 if $A$
is a tautology, then $Pr(A) = 1$. }

We also extend this definition for conditional probability. We again
introduce a new operator $Pr$ over pairs of two assertions.
Following a similar idea, the conditional probability $Pr(A_1,A_2)$
of an assertion $A_1$ providing another assertion $A_2$, also
denoted by $Pr(A_1|A_2)$, is defined by the following schema
assertion:
\begin{equation}\label{def-cond-probability}
Pr(A_1|A_2)= \frac{\Sigma_{w, w \models A_1, w \models A_2} \:\:
W_w}{\Sigma_{w, w \models A_2} \: \:W_w}.
\end{equation}

Again, once conditional probability is defined as a real number, we
can use it arbitrarily inside other assertions. Similarly, we can
verify some properties about conditional probabilities, including
the famous Bayes' theorem, i.e.,
\[
Pr(A_1)\times Pr(A_2|A_1) = Pr(A_2)\times Pr(A_1|A_2).
\]
for all assertions $A_1$ and $A_2$.

\subsection{Interactions between logic and probability via assertions}

Although we only define probabilities for assertions of the basic
form, the interactions between probability and other building
blocks, e.g., logic, are automatically established since assertions
connected by logic operators can be reduced into the primitive form.
In this sense, we can investigate some properties about the
interactions between logic and probability. For instance, it can be
observed that Kolmogorov's third probability axiom is a tautology in
assertion logic. That is, let $A_1,\dots,A_n$ be $n$ assertions that
are pairwise disjoint. Then, $Pr(A_1 \lor \dots \lor A_n) = Pr(A_1)+
\dots + Pr(A_n)$.

\comment{\footnote{Note that Kolmogorov's probability axioms are
defined for events instead of assertions. To represent an event in
assertional logic, one can use its effects (i.e., postconditions),
which is an assertion. Then, that two events are disjoint if and
only if their postconditions do not have common models.}.}

It can be verified that many axioms and properties regarding the
interactions between logic and probability are tautologies in
assertional logic, for instance, the additivity axiom:
$Pr(\phi)=Pr(\phi\land \psi) + Pr(\phi \land \neg \psi)$ and the
distributivity axiom: $\phi \equiv \psi$ implies that
$Pr(\phi)=Pr(\psi)$, for any two assertions $\phi$ and $\psi$. In
this sense,  assertional logic can also be used to validate existing
properties about the interactions of logic and probability. In
addition, it may foster new discoveries, e.g., the interactions
between higher-order logic and probability and some properties about
nested probabilities.

Note that we do not intend to reinvent the wheel of defining
probability nor its interactions with logic. All definitions about
(conditional) probability are borrowed from the literature. Instead,
we take probability as a case study to show how one building block
(e.g., logic) and another (e.g., probability) can be interacted
through assertions without going deeper into the interactions
between themselves.

\subsection{More building blocks}

More critically, there are many more important building blocks to be
incorporated. It is barely possible to clarify the interactions
among them all. Nevertheless, it becomes possible to unify them
altogether in assertional logic as one only needs to consider the
interactions between these building blocks and the basic form of
assertions separately. Consequently, the interactions among these
building blocks themselves will be automatically established via
assertions, as what we did for unifying logic and probability.

As another case study, we consider how to formalize time in
assertional logic. Time itself can be understood in different ways
such as time points, time interval, LTL and CTL
\cite{Allen1983,Clarke1981,Pnueli1977}. Following the same idea of
incorporating probability, we only need to consider the interactions
between time and assertions. In this paper, we only report the
simple case of integrating assertions with time points. Let $Tp$ be
a concept of time points. We introduce a new operator $\mathtt{t}$
whose domain is a pair $(\mathcal{I}, Tp)$. Intuitively,
$\mathtt{t}(i,tp), i \in \mathcal{T}, tp \in Tp$, is the value of
individual $i$ at time point $tp$. Then, we introduce temporal
formulas, a new Boolean operator $\mathcal{T}$ whose domain is a
pair $(\mathcal{A}, Tp)$, by the following schema assertion:
\begin{equation}
\mathcal{T}(a=b,tp) = \mathtt{t}(a,tp)=\mathtt{t}(b,tp).
\end{equation}
Then, the interactions between time points and logic connectives and
probability can be automatically established. We are able to
investigate some properties. For instance, for all assertions $A$
and $B$, $\mathcal{T}(A,tp) = \top$ iff $\mathcal{T}(\lnot
A,tp)=\bot$; $\mathcal{T}(A \land B,tp)=\top \models
\mathcal{T}(A,tp) =\top$ etc. Hence, we have an integrated formalism
combing logic, probability and time points in assertional logic.

\comment{ One can further introduce new concepts and operators for
other temporal assertions, e.g., time interval, and define
relationships among time intervals similar to Allen's interval
algebra \cite{Allen1983} Again, the key point is that for extending
with a new building block (e.g., time interval) in assertional
logic, one only needs to formalize it by syntactic objects and
define its interactions with the basic form (\ref{form-assertion})
of assertions. Then, the interactions between this building block
and others (e.g., logic, probability or time points) will be
automatically established through assertions. Moreover, one can
freely assembly some of the building blocks together as different
fragments in assertional logic.}

\section{Discussion, Related Work and Conclusion}

In this paper, we argue that, for the purpose of knowledge
representation, classical first-order logic has some critical
issues, including simplicity, flexibility, expressivity and
extensibility. To address these issues, we propose assertional logic
instead, in which the syntax of an application domain is captured by
individuals (i.e., objects in the domain), concepts (i.e., groups of
objects sharing something in common) and operators (i.e.,
connections and relationships among objects), and knowledge in the
domain is simply captured by equality assertions of the form $a=b$,
where $a$ and $b$ are terms.

In assertional logic, without redefining the semantics, one can
extend a current system with new syntactic objects by definitions,
which are special kinds of knowledge (i.e., assertions). Once
defined, these syntactic objects can be used to define more. This
can be done for assertional logic itself. We extend the primitive
form of assertional logic with multi-assertions and nested
assertions as well as logic connectives and quantifiers. We further
consider how to extend assertional logic with other important AI
building blocks. The key point is that, when one wants to integrate
a new building block in assertional logic, she only needs to
formalize it as syntactic objects (including individuals, concepts
and operators) and defines its interactions with the basic form of
assertions (i.e., $a=b$). Then, the interactions between this
building block and others will be automatically established since
all complicated assertions can essentially be reduced into the basic
form. As a case study, we briefly discuss how to incorporate
probability and time points in this paper.

Of course, assertional logic is deeply rooted in first-order logic.
Individuals, concepts and operators are analogous to constants,
unary predicates and functions respectively, and assertions are
originated from equality atoms. Nevertheless, they differ from many
essential aspects. Firstly, individuals can be high-order objects,
e.g., concepts and assertions, so are concepts and operators.
Secondly, assertional logic is naturally many-sorted, that is, the
domain of an operator can be a tuple of many different concepts
including high-order ones. Thirdly, concepts play a central role in
assertional logic, which is natural for human knowledge
representation. While concepts can be formalized as unary predicates
in FOL, they are not specifically emphasized. Fourthly, in
assertional logic, all kinds of knowledge are uniformly formalized
in the same form of equality assertions. As shown in Section 5,
complicated logic sentences are defined as equality assertions as
well by embedding connectives and quantifiers as operators over
assertions. Fifthly, following the above, although connectives,
quantifiers and nesting can be defined in assertional logic, they
are not considered as primitive constructs. In this sense, they will
only be used on demand when necessary. We argue that this is an
important reason that makes assertional logic simpler than FOL.
Sixthly, in assertional logic, the simple form of $a=b$ is
expressive as $a$ and $b$ can be high-order constructs and can be
inherently related within the rich syntactic structure. In contrast,
equality atoms in FOL do not have this power. Last but not least,
assertional logic directly embraces extensibility within its own
framework by syntactic definitions. For instance, to define
quantifiers, assertional logic only needs two lines (see Equations
\ref{universal-quantifier} and \ref{exist-quantifier}) without
redefining a whole new semantics, which is much simpler than FOL.

Assertional logic is also inspired by many approaches in symbolic
AI. Traditionally in FOL, there is a strict hierarchy from the term
level to the formula level. To some extent, this is broken in
situation calculus \cite{levesque1991,Lin08} and game description
language \cite{Thielscher16} that have to quantify over situations,
actions and fluents and to directly talk about whether a fluent
holds in a particular situation by a meta-predicate $Hold$.
Assertional logic goes much further by completely demolishing the
distinction and hierarchy between term level, predicate level and
formula level. In assertional logic, one can flexibly use, e.g.,
atoms and predicates in the scope of a function as long as they
match its domain definition. Also, one can quantify over any
well-defined concepts, including a concept of assertions. This makes
assertional logic even more expressive than high-order logic that
cannot quantifier over formulas.

Another inspiration comes from the family of description logics
\cite{Baader03}, where the terminologies individuals and concepts
are borrowed from. The family of description logics allows a certain
level of extensibility. By interpreting individuals, concepts and
roles as domain elements, unary predicates and binary predicates
respectively, one can extend the basic description logics with more
building blocks (e.g., nominal, number restrictions, inverse and
transitive roles etc.) based on the same foundational semantics.
Also, one can freely assemble those building blocks into different
fragments of description logics such as ALC, SHIQ, SHION and so on.
However, many important AI building blocks, e.g., actions,
probability, time, rules, etc. are still difficult to be
incorporated by this method. Some interesting pioneering work have
been done to consider more extensibility in description logics
\cite{Baader91,Borgida99,Giacomo11,KuLuWoZa-AI-03}. Nevertheless,
they differ from assertional logic that embraces extensibility at a
syntactic level instead of a semantic one.

This paper is only concerned with the representation task and the
definition task. We leave the reasoning task to our future work.
Certainly, complete reasoning in assertional logic is undecidable as
FOL can be embedded in it. Traditionally, the way to address this
issue is to find decidable fragments. Nevertheless, we shall propose
an alternative approach that focuses on efficient but not
necessarily complete reasoning. We have developed such an approach,
in which the flexibility and extensibility of assertional logic play
a critical role. We shall present this in another paper.
Nevertheless, we argue that representation and definition are worth
study on their own merits. Such successful stories include
entity-relationship diagram, semantic network and many more.
Besides, extending assertional logic with some important AI building
blocks, e.g., actions and their effects, is indeed challenging and
worth pursuing.

\section*{Acknowledgement}

The author gratefully acknowledges Prof. Fangzhen Lin's comments on
a first draft of this paper.

\bibliographystyle{plain}
\bibliography{set-kr}

\begin{thebibliography}{10}

\bibitem{Allen1983}
James~F. Allen.
\newblock Maintaining knowledge about temporal intervals.
\newblock {\em Commun. ACM}, 26(11):832--843, November 1983.

\bibitem{Baader03}
Franz Baader, Diego Calvanese, Deborah~L. McGuinness, Daniele Nardi, and
  Peter~F. Patel-Schneider, editors.
\newblock {\em The Description Logic Handbook: Theory, Implementation, and
  Applications}.
\newblock Cambridge University Press, New York, NY, USA, 2003.

\bibitem{Baader91}
Franz Baader and Philipp Hanschke.
\newblock A scheme for integrating concrete domains into concept languages.
\newblock In {\em Proceedings of the 12th International Joint Conference on
  Artificial Intelligence - Volume 1}, IJCAI'91, pages 452--457, San Francisco,
  CA, USA, 1991. Morgan Kaufmann Publishers Inc.

\bibitem{Bacchus1991}
Fahiem Bacchus.
\newblock {\em Representing and Reasoning with Probabilistic Knowledge: A
  Logical Approach to Probabilities}.
\newblock MIT Press, Cambridge, MA, USA, 1990.

\bibitem{Borgida99}
Alexander Borgida.
\newblock Extensible knowledge representation: the case of description
  reasoners.
\newblock {\em J. Artif. Intell. Res. {(JAIR)}}, 10:399--434, 1999.

\bibitem{Brachman04}
Ronald~J. Brachman and Hector~J. Levesque.
\newblock {\em Knowledge Representation and Reasoning}.
\newblock Elsevier, 2004.

\bibitem{Clarke1981}
Edmund~M. Clarke and E.~Allen Emerson.
\newblock Design and synthesis of synchronization skeletons using
  branching-time temporal logic.
\newblock In {\em Logic of Programs, Workshop}, pages 52--71, London, UK, UK,
  1982. Springer-Verlag.

\bibitem{Gaifman64}
Haim Gaifman.
\newblock Concerning measures in first order calculi.
\newblock {\em Israel J. Math.}, 2:1--18, 1964.

\bibitem{Giacomo11}
Giuseppe~De Giacomo, Maurizio Lenzerini, and Riccardo Rosati.
\newblock Higher-order description logics for domain metamodeling.
\newblock In {\em Proceedings of the Twenty-Fifth AAAI Conference on Artificial
  Intelligence}, AAAI'11, pages 183--188. AAAI Press, 2011.

\bibitem{zbMATH03871328}
Theodore {Hailperin}.
\newblock {Probability logic.}
\newblock {\em {Notre Dame J. Formal Logic}}, 25:198--212, 1984.

\bibitem{Halmos60}
Paul Halmos.
\newblock {\em Naive Set Theory}.
\newblock Van Nostrand, 1960.
\newblock Reprinted by Springer-Verlag, Undergraduate Texts in Mathematics,
  1974.

\bibitem{zbMATH04191596}
Joseph~Y. {Halpern}.
\newblock {An analysis of first-order logics of probability.}
\newblock {\em {Artif. Intell.}}, 46(3):311--350, 1990.

\bibitem{Keller10}
Thomas Keller, Patrick Eyerich, and Bernhard Nebel.
\newblock Task planning for an autonomous service robot.
\newblock In {\em Proceedings of the 33rd Annual German Conference on Advances
  in Artificial Intelligence}, KI'10, pages 358--365, Berlin, Heidelberg, 2010.
  Springer-Verlag.

\bibitem{KuLuWoZa-AI-03}
O.~{Kutz}, C.~{Lutz}, F.~{Wolter}, and M.~{Zakharyaschev}.
\newblock E-connections of abstract description systems.
\newblock {\em Artificial Intelligence}, 156(1):1--73, 2004.

\bibitem{levesque1991}
Hector Levesque, Fiora Pirri, and Ray Reiter.
\newblock Foundations for the situation calculus.
\newblock {\em Electronic Transactions on Artificial Intelligence}, Vol.
  2(1998), Issue 3-4:pp. 159--178, 1991.

\bibitem{Lin08}
Fangzhen Lin.
\newblock Situation calculus.
\newblock In {\em Handbook of Knowledge Representation}, pages 649--669. 2008.

\bibitem{Milch:2006:PMU:1292992}
Brian~Christopher Milch.
\newblock {\em Probabilistic Models with Unknown Objects}.
\newblock PhD thesis, Berkeley, CA, USA, 2006.
\newblock AAI3253991.

\bibitem{Pearl1988}
Judea Pearl.
\newblock {\em Probabilistic Reasoning in Intelligent Systems: Networks of
  Plausible Inference}.
\newblock Morgan Kaufmann Publishers Inc., San Francisco, CA, USA, 1988.

\bibitem{Pnueli1977}
Amir Pnueli.
\newblock The temporal logic of programs.
\newblock In {\em Proceedings of the 18th Annual Symposium on Foundations of
  Computer Science}, SFCS '77, pages 46--57, Washington, DC, USA, 1977. IEEE
  Computer Society.

\bibitem{Richardson06}
Matthew Richardson and Pedro Domingos.
\newblock Markov logic networks.
\newblock {\em Machine learning}, 62(1-2):107--136, 2006.

\bibitem{Thielscher16}
Michael Thielscher.
\newblock {GDL-III:} {A} proposal to extend the game description language to
  general epistemic games.
\newblock In {\em {ECAI} 2016 - 22nd European Conference on Artificial
  Intelligence, 29 August-2 September 2016, The Hague, The Netherlands -
  Including Prestigious Applications of Artificial Intelligence {(PAIS} 2016)},
  pages 1630--1631, 2016.

\bibitem{Harmelen08}
Frank van Harmelen, Vladimir Lifschitz, and Bruce~W. Porter, editors.
\newblock {\em Handbook of Knowledge Representation}, volume~3 of {\em
  Foundations of Artificial Intelligence}.
\newblock Elsevier, 2008.

\end{thebibliography}

\end{document}